\documentclass{article}%
\usepackage{amsmath}
\usepackage{amsfonts}
\usepackage{amssymb}
\usepackage{graphicx}
\usepackage{algorithmic}
\usepackage{algorithm}%
\providecommand{\U}[1]{\protect\rule{.1in}{.1in}}

\begin{document}

\title{Heterogeneous Treatment Effect with Trained Kernels of the Nadaraya-Watson Regression}
\author{Andrei V. Konstantinov, Stanislav R. Kirpichenko, Lev V. Utkin\\Peter the Great St.Petersburg Polytechnic University\\St.Petersburg, Russia\\e-mail: andrue.konst@gmail.com, kirpichenko.sr@gmail.com, lev.utkin@gmail.com}
\date{}
\maketitle

\begin{abstract}
A new method for estimating the conditional average treatment effect is
proposed in the paper. It is called TNW-CATE (the Trainable Nadaraya-Watson
regression for CATE) and based on the assumption that the number of controls
is rather large whereas the number of treatments is small. TNW-CATE uses the
Nadaraya-Watson regression for predicting outcomes of patients from the
control and treatment groups. The main idea behind TNW-CATE is to train
kernels of the Nadaraya-Watson regression by using a weight sharing neural
network of a specific form. The network is trained on controls, and it
replaces standard kernels with a set of neural subnetworks with shared
parameters such that every subnetwork implements the trainable kernel, but the
whole network implements the Nadaraya-Watson estimator. The network memorizes
how the feature vectors are located in the feature space. The proposed
approach is similar to the transfer learning when domains of source and target
data are similar, but tasks are different. Various numerical simulation
experiments illustrate TNW-CATE and compare it with the well-known T-learner,
S-learner and X-learner for several types of the control and treatment outcome
functions. The code of proposed algorithms implementing TNW-CATE is available
in https://github.com/Stasychbr/TNW-CATE.

\textit{Keywords}: treatment effect, Nadaraya-Watson regression, neural
network, shared weights, meta-learner, regression.

\end{abstract}

\section{Introduction}

The efficient treatment for a patient with her/his clinical and other
characteristics \cite{Lu-Sadiq-etal-2017,Shalit-etal-2017} can be regarded as
an important goal of the real personalized medicine. The goal can be achieved
by means of the machine learning methods due to the increasing amount of
available electronic health records which are a basis for developing accurate
models. To estimate the treatment effect, patients are divided into two groups
called treatment and control, and then patients from the different groups are
compared. One of the popular measures of the efficient treatment used in
machine learning models is the average treatment effect (ATE)
\cite{Fan-Lv-Wang-2018}, which is estimated on the basis of observed data
about patients as the mean difference between outcomes of patients from the
treatment and control groups. Due to the difference between the patients
characteristics and the difference between their responses to a particular
treatment, the treatment effect is measured by the conditional average
treatment effects (CATE) or the heterogeneous treatment effect (HTE) defined
as ATE conditional on a patient feature vector
\cite{Green-Kern-2012,Hill-2011,Kallus-2016,Wager-Athey-2015}.

Two main problems can be pointed out when CATE is estimated. The first one is
that the control group is usually larger than the treatment group. As a
result, we meet the problem of a small training dataset, which does not allow
us to apply directly many efficient machine learning methods. The second
problem is that each patient cannot be simultaneously in the treatment and
control groups, i.e., we either observe the patient outcome under treatment or
control, but never both \cite{Kunzel-etal-2018a}. This is a fundamental
problem of computing the causal effect. In addition to the above two problems,
there are many difficulties of the machine learning model development
concerning with noisy data, with the high dimension of the patient health
records, etc. \cite{Wendling-etal-2018}.

Many methods for estimating CATE have been proposed and developed due to
importance of the problem in medicine and other applied areas
\cite{Acharki-etal-22,Alaa-Schaar-2018,Athey-Imbens-2016,Deng-etal-2016,Fernandez-Loria-Provost-22,Fernandez-Loria-Provost-22a,Gong-Hu-etal-21,Hatt-etal-22,Jiang-Qi-etal-21,Kunzel-etal-2018,Kunzel-etal-19,Utkin-Kots-etal-2020,Wu-Yang-22,Yadlowsky-etal-21,Zhang-Li-Liu-22}%
. This is only a small part of all publications which are devoted to solving
the problem of estimating CATE. Various approaches were used for solving the
problem, including the support vector machine \cite{Zhao-etal-2012},
tree-based models \cite{Wendling-etal-2018}, neural networks
\cite{Bica-etal-20,Curth-Schaar-21a,Kunzel-etal-2018a,Shalit-etal-17},
transformers
\cite{Guo-Zheng-etal-21,Melnychuk-etal-22,Zhang-Zhang-etal-22,Zhang-Zhang-etal-22a}%
.

It should be noted that most approaches to estimating CATE are based on
constructing regression models for handling the treatment and control groups.
However, the problem of the small treatment group motivates to develop various
tricks which at least partially resolve the problem.

We propose a method based on using the Nadaraya-Watson kernel regression
\cite{Nadaraya-1964,Watson-1964} which is widely applied to machine learning
problems. The method is called TNW-CATE (the Trainable Nadaraya-Watson
regression for CATE). The Nadaraya-Watson estimator can be seen as a weighted
average of outcomes (patient responses) by means of a set of varying weights
called attention weights. The attention weights in the Nadaraya-Watson
regression are defined through kernels which measure distances between
training feature vectors and the target feature vector, i.e., kernels in the
Nadaraya-Watson regression conform with relevance of a training feature vector
to a target feature vector. If we have a dataset $\{(\mathbf{x}_{1}%
,y_{1}),...,(\mathbf{x}_{n},y_{n})\}$, where $\mathbf{x}_{i}\in\mathbb{R}^{m}$
is a feature vector (key) and $y_{i}\in\mathbb{R}$ is its target value or its
label (value), then the Nadaraya-Watson estimator for a target feature vector
$\mathbf{z}\in\mathbb{R}^{m}$ (query) can be defined by using weights
$\alpha(\mathbf{z},\mathbf{x}_{i})$ as
\begin{equation}
\hat{f}(\mathbf{z})=\sum_{i=1}^{n}\alpha(\mathbf{z},\mathbf{x}_{i}%
)y_{i},\ \alpha(\mathbf{z},\mathbf{x}_{i})=\frac{K(\mathbf{z},\mathbf{x}%
_{i},\gamma)}{\sum_{j=1}^{n}K(\mathbf{z},\mathbf{x}_{j},\gamma)},
\end{equation}
where $K(\mathbf{z},\mathbf{x}_{i},\gamma)$ is a kernel and $\gamma>0$ is a
bandwidth parameter.

Standard kernels widely used in practice are the Gaussian, uniform, or
Epanechnikov kernels \cite{Bartlett-etal-21}. However, the choice of a kernel
and its parameters significantly impact on results obtained from the
Nadaraya-Watson regression usage. Moreover, the Nadaraya-Watson regression
also requires a large number of training examples. Therefore, we propose a
quite different way for implementing the Nadaraya-Watson regression. The way
is based on the following assumptions and ideas. First, each kernel is
represented as a part of a neural network implementing the Nadaraya-Watson
regression. In other words, we do not use any predefined standard kernels like
the Gaussian ones. Kernels are trained as the weight sharing neural
subnetworks. The weight sharing is used to identically compute kernels under
condition that pair $(\mathbf{z},\mathbf{x}_{i})$ of examples is fed into
every subnetwork. The neural network kernels become more flexible and
sensitive to a complex location structure of feature vectors. In fact, we
propose to replace the definition of weights through kernels with a set of
neural subnetworks with shared parameters (the neural network weights) such
that every subnetwork implements the trainable kernel, but the whole network
implements the Nadaraya-Watson estimator. At that, the trainable parameters of
kernels are nothing else but weights of each neural subnetwork. The above
implementation of the Nadaraya-Watson regression by means of the neural
network leads to an interesting result when the treatment examples are
considered as a single example whose \textquotedblleft
features\textquotedblright\ are the whole treatment feature vectors.

\begin{figure}
[ptb]
\begin{center}
\includegraphics[
height=1.8917in,
width=4.2206in
]%
{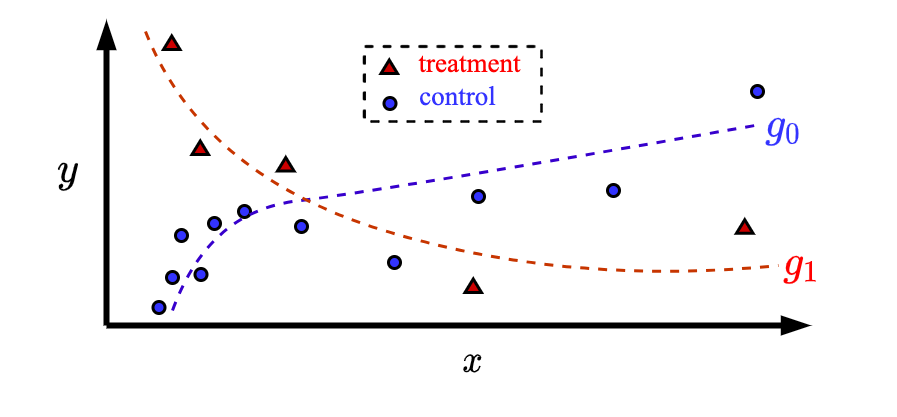}%
\caption{Illustration of the logarithmical location of feature vectors
corresponding to patients from the treatment group (small triangles) and from
the control group (small circles)}%
\label{f:hte_1}%
\end{center}
\end{figure}

\begin{figure}
[ptb]
\begin{center}
\includegraphics[
height=3.3121in,
width=3.7404in
]%
{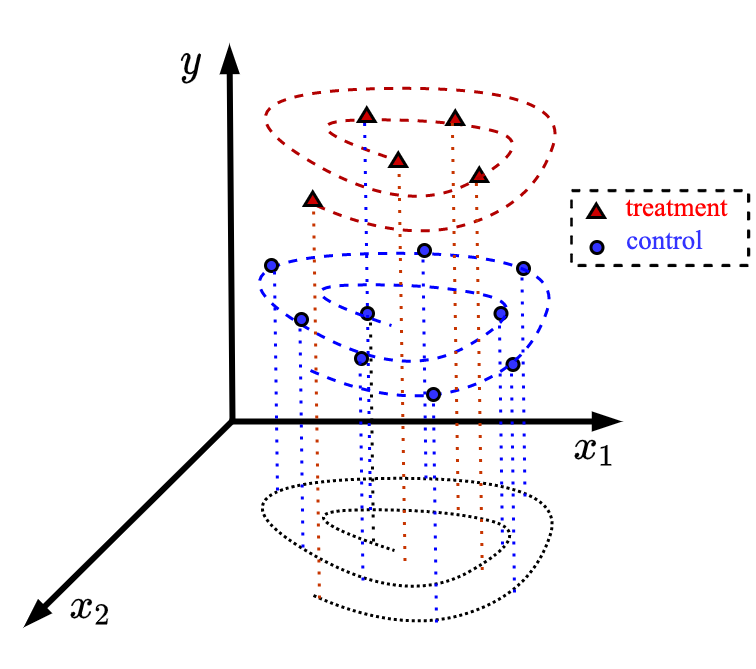}%
\caption{Illustration of the location of feature vectors corresponding to
patients from the treatment group (small triangles) and from the control group
(small circles) on spirals}%
\label{f:hte_2}%
\end{center}
\end{figure}

Second, it is assumed that the feature vector domains of the treatment and
control groups are similar. For instance, if some components of the feature
vectors from the control group are logarithmically located in the feature
space, then the feature vectors from the treatment group have the same
tendency to be located in the feature space. Fig. \ref{f:hte_1} illustrates
the corresponding location of the feature vectors. Vectors from the control
and treatment groups are depicted by small circles and triangles,
respectively. It can be seen from Fig. \ref{f:hte_1} that the control examples
as well as the treatment ones are located unevenly along the x-axis. Many
standard regression methods do not take into account this peculiarity. Another
illustration is shown in Fig. \ref{f:hte_2} where the feature vectors are
located on spirals, but the spirals have different values of the patient
outcomes. Even if we were to use a kernel regression, it is difficult to find
such standard kernels that would satisfy training data. However, the
assumption of similarity of domains allows us to train kernels on examples
from the control group because kernels depend only on the feature vectors. In
this case, the network memorizes how the feature vectors are located in the
feature space. By using assumption about similarity of the treatment and
control domains, we can apply the Nadaraya-Watson regression with the trained
kernels to the treatment group changing the patient outcomes. TNW-CATE is
similar to the transfer learning
\cite{Lu-etal-2015,Pan-Yang-2010,Weiss-etal-2016} when domains of source and
target data are the same, but tasks are different. Therefore, the abbreviation
TNW-CATE can be also read as the Transferable kernels of the Nadaraya-Watson
regression for the CATE estimating.%

%

Various numerical experiments study peculiarities of TNW-CATE and illustrate
the outperformance of TNW-CATE in comparison with the well-known meta-models:
the T-learner, the S-learner, the X-learner for several control and treatment
output functions. The code of the proposed algorithms can be found in https://github.com/Stasychbr/TNW-CATE.

The paper is organized as follows. Section 2 can be viewed as a review of
existing models for estimating CATE. Applications of the Nadaraya-Watson
regression in machine learning also are considered in this section. A formal
statement of the CATE estimation problem is provided in Section 3. TNW-CATE,
its main ideas and algorithms implementing TNW-CATE are described in Section
4. Numerical experiments illustrating TNW-CATE and comparing it with other
models are presented in Section 5. Concluding remarks are provided in Section 6.

\section{Related work}

\textbf{Estimating CATE}. Computing CATE is a very important problem which can
be regarded as a tool for implementing the personalized medicine
\cite{Powers-etal-2017}. This fact motivated to develop many efficient
approaches solving the problem. An approach based on the Lasso model for
estimating CATE was proposed by Jeng et al. \cite{Jeng-Lu-Peng-2018}. Several
interesting approaches using the SVM model were presented in
\cite{Zhao-etal-2012,Zhou-Mayer-Hamblett-etal-2017}. A \textquotedblleft
honest\textquotedblright\ model for computing CATE was proposed by Athey and
Imbens \cite{Athey-Imbens-2016}. According to the model, the training set is
splitting into two subsets such that the first one is used to construct the
partition of the data into subpopulations that differ in the magnitude of
their treatment effect, and the second subset is used to estimate treatment
effects for each subpopulation. A unified framework for constructing fast
tree-growing procedures solving the CATE problem was provided in
\cite{Athey-Tibshirani-Wager-2016,Athey-Tibshirani-Wager-2018}. A modification
of the survival causal tree method for estimating the CATE based on censored
observational data was proposed in \cite{Zhang-Le-etal-2017}. Xie et al.
\cite{Xie-Chen-Shi-2018} established the CATE detection problem as a false
positive rate control problem, and discussed in details the importance of this
approach for solving large-scale CATE detection problems. Algorithms for
estimating CATE in the context of Bayesian nonparametric inference were
studied in \cite{Alaa-Schaar-2018}. Bayesian additive regression trees, causal
forest, and causal boosting were compared under condition of binary outcomes
in \cite{Wendling-etal-2018}. An orthogonal random forest as an algorithm that
combines orthogonalization with generalized random forests for solving the
CATE estimation problem was proposed in \cite{Oprescu-Syrgkanis-Wu-2018}.
Estimating CATE as the anomaly detection problem was studied in
\cite{McFowland-etal-2018}. Many other approaches can also be found in
\cite{Chen-Liu-2018,Grimmer-etal-017,Kallus-Puli-Shalit-2018,Kallus-Zhou-2018,Knaus-etal-2018,Kunzel-etal-2018b,Levy-2018,Powers-etal-2017,Rhodes-2010,Xie-Brand-Jann-2012,Yao-Lo-Nir-etal-22}%
.

A set of meta-algorithms or meta-learners, including the T-learner
\cite{Kunzel-etal-2018}, the S-learner \cite{Kunzel-etal-2018}, the O-learner
\cite{Wang-etal-2016}, the X-learner \cite{Kunzel-etal-2018} were investigated
and compared in \cite{Kunzel-etal-2018}.

Neural networks can be regarded as one of the efficient tools for estimating
CATE. As a result, many models based on using neural networks have been
proposed
\cite{Bica-etal-20,Curth-Schaar-21,Curth-Schaar-21a,Du-Fan-etal-21,Kunzel-etal-2018a,Nair-etal-22,Nie-etal-21,Parbhoo-etal-21,Qin-Wang-Zhou-21,Schwab-etal-2020,Shalit-etal-17,Shi-Blei-Veitch-19,Veitch-etal-19}%
.

Transformer-based architectures using attention operations
\cite{Chaudhari-etal-2019} were also applied to solving the CATE estimating
problem
\cite{Guo-Zheng-etal-21,Melnychuk-etal-22,Zhang-Zhang-etal-22,Zhang-Zhang-etal-22a}%
. Ideas of applying the transfer learning technique to the CATE estimation
were considered in
\cite{Aoki-Ester-22,Guo-Wang-etal-21,Kunzel-etal-18,Kunzel-etal-2018a}. Ideas
of using the Nadaraya-Watson kernel regression in the CATE estimation were
studied in \cite{Imbens-04,Park-Shalit-etal-21} where it was shown that the
Nadaraya-Watson regression can be used for the CATE estimation. However, the
small number of training examples in the treatment group does not allow us to
efficiently apply this approach. Therefore, we aim to overcome this problem by
introducing a neural network of a special architecture, which implements the
trainable kernels in the Nadaraya-Watson regression.

\textbf{The Nadaraya-Watson regression in machine learning}. There are several
machine learning approaches based on applying the Nadaraya-Watson regression
\cite{Ghassabeh-Rudzicz-16,Hanafusa-Okadome-20,Konstantinov-Utkin-22d,Liu-Huang-etal-21a,Shapiai-etal-10,Xiao-Xiang-etal-19,Zhang-14}%
. Properties of the boosting with kernel regression estimates as weak learners
were studied in \cite{Park-Lee-Ha-09}. A metric learning model with the
Nadaraya-Watson kernel regression was proposed in \cite{Noh-etal-17}. The
high-dimensional nonparametric regression models were considered in
\cite{Conn-Li-19}. Models taking into account available correlated errors were
proposed in \cite{DeBrabanter-etal-11}. An interesting work discussing a
problem of embedding the Nadaraya-Watson regression into the neural network as
a novel trainable CNN layer was presented in \cite{Szczotka-etal-20}. Applied
machine learning problems solved by using the Nadaraya-Watson regression were
considered in \cite{Liu-Min-etal-21}. A method of approximation using the
kernel functions made from only the sample points in the neighborhood of input
values to simplify the Nadaraya-Watson estimator is proposed in
\cite{Ito-etal-20}. An interesting application of the Nadaraya-Watson
regression to improving the local explanation method SHAP is presented in
\cite{Ghalebikesabi-etal-21} where authors find that the Nadaraya-Watson
estimator can be expressed as a self-normalized importance sampling estimator.
An explanation of how the Nadaraya-Watson regression can be regarded as a
basis for understanding the attention mechanism from the statistics point of
view can be found in \cite{Chaudhari-etal-2019,Zhang2021dive}.

In contrast to the above works, we pursue two goals. The first one is to show
how kernels in the Nadaraya-Watson regression can be implemented and trained
as neural networks of a special form. The second goal is to apply the whole
neural network implementing the Nadaraya-Watson regression to the problem of
estimating CATE.

\section{A formal problem statement}

Suppose that the control group of patients is represented as a set of $c$
examples of the form $\mathcal{C}=\{(\mathbf{x}_{1},y_{1}),...,(\mathbf{x}%
_{c},y_{c})\}$, where $\mathbf{x}_{i}=(x_{i1},...,x_{id})\in\mathbb{R}^{d}$ is
the $d$-dimensional feature vector for the $i$-th patient from the control
group; $y_{i}\in\mathbb{R}$ is the $i$-th observed outcome, for example, time
to death of the $i$-th patient from the control group or the blood sugar level
of this patient. The similar notations can be introduced for the treatment
group containing $t$ patients, namely, $\mathcal{T}=\{(\mathbf{z}_{1}%
,h_{1}),...,(\mathbf{z}_{t},h_{t})\}$. Here $\mathbf{z}_{i}=(z_{i1}%
,...,z_{id})\in\mathbb{R}^{d}$ and $h_{i}\in\mathbb{R}$ are the feature vector
and the outcome of the $i$-th patient from the treatment group, respectively.
We will also use the notation of the treatment assignment indicator $T_{i}%
\in\{0,1\}$ where $T_{i}=0$ ($T_{i}=1$) corresponds to the control (treatment) group.

Let $Y$ and $H$ denote the potential outcomes of a patient if he had received
treatment $T=0$ and $T=1$, respectively. Let $\mathbf{X}$ be the random
feature vector from $\mathbb{R}^{d}$. The treatment effect for a new patient
with the feature vector $\mathbf{x}$, which shows how the treatment is useful
and efficient, is estimated by the Individual Treatment Effect (ITE) defined
as $H-Y$. Since the ITE cannot be observed, then the treatment effect is
estimated by means of CATE which is defined as the expected difference between
two potential outcomes as follows \cite{Rubin-2005}:
\begin{equation}
\tau(\mathbf{x})=\mathbb{E}\left[  H-Y|\mathbf{X}=\mathbf{x}\right]  .
\end{equation}

The fundamental problem of computing CATE is that, for each patient in the
training dataset, we can observe only one of outcomes $y$ or $h$ for each
patient. To overcome this problem, the important assumption of
unconfoundedness \cite{Rosenbaum-Rubin-1983} is used to allow the untreated
units to be used to construct an unbiased counterfactual for the treatment
group \cite{Imbens-2004}. According to the assumption of unconfoundedness, the
treatment assignment $T$ is independent of the potential outcomes for $Y$ or H
conditional on $\mathbf{x}=\mathbf{z}$, respectively, which can be written as
\begin{equation}
T\perp\{Y,H\}|\mathbf{x}.
\end{equation}

Another assumption called the overlap assumption regards the joint
distribution of treatments and covariates. According to the assumption, there
is a positive probability of being both treated and untreated for each value
of $\mathbf{x}$. It is of the form:
\begin{equation}
0<\Pr\{T=1|\mathbf{x}\}<1.
\end{equation}

If to accept these assumptions, then CATE\ can be represented as follows:
\begin{equation}
\tau(\mathbf{x})=\mathbb{E}\left[  H|\mathbf{X}=\mathbf{x}\right]
-\mathbb{E}\left[  Y|\mathbf{X}=\mathbf{z}\right]  .
\end{equation}

The motivation behind unconfoundedness is that nearby observations in the
feature space can be treated as having come from a randomized experiment
\cite{Wager-Athey-2017}.

If we suppose that outcomes of patients from the control and treatment groups
are expressed through the functions $g_{0}$ and $g_{1}$ of the feature vectors
$X$, then the corresponding regression functions can be written as
\begin{equation}
y=g_{0}(\mathbf{x})+\varepsilon,\ \mathbf{x}\in\mathcal{C},
\end{equation}%
\begin{equation}
h=g_{1}(\mathbf{z})+\varepsilon,\ \mathbf{z}\in\mathcal{T}.
\end{equation}

Here $\varepsilon$ is a is a Gaussian noise variable such that $\mathbb{E}%
(\varepsilon)=0$. Hence, there holds under condition $\mathbf{x}=\mathbf{z}$
\begin{equation}
\tau(\mathbf{x})=g_{1}(\mathbf{x})-g_{0}(\mathbf{x}). \label{HTE_concat_20}%
\end{equation}

An example illustrating sets of controls (small circles), treatments (small
triangles) and the corresponding unknown function $g_{0}$ and $g_{1}$ are
shown Fig. \ref{f:hte_1}.

\section{The TNW-CATE description}

It has been mentioned that the main idea behind TNW-CATE is to replace the
Nadaraya-Watson regression with the neural network of a specific form. The
whole network consists of two main parts. The first part implements the
Nadaraya-Watson regression for training the control function $g_{0}%
(\mathbf{x})$. In turn, it consists of $n$ identical subnetworks such that
each subnetwork implements the attention weight $\alpha(\mathbf{x}%
,\mathbf{x}_{i})$ or the kernel $K(\mathbf{x},\mathbf{x}_{i})$ of the
Nadaraya-Watson regression. Therefore, the input of each subnetwork is two
vectors $\mathbf{x}$ and $\mathbf{x}_{i}$, i.e., two vectors $\mathbf{x}$ and
$\mathbf{x}_{i}$ are fed to each subnetwork. The whole network consisting of
$n$ identical subnetworks and implementing the Nadaraya-Watson regression for
training the control function will be called the control network. In order to
train the control network, for every vector $\mathbf{x}_{i}$, $i=1,...,c$, $N$
subsets of size $n$ are randomly selected from the control set $\mathcal{C}$
without example $(\mathbf{x}_{i},y_{i})$. The subsets can be regarded as $N$
examples for training the network. Hence, the control network is trained on
$N\cdot c$ examples of size $2d\cdot n$. If we have a feature vector
$\mathbf{x}$ for estimating $\tilde{y},$ i.e., for estimating function
$g_{0}(\mathbf{x})$, then it is fed to each trained subnetworks jointly with
each $\mathbf{x}_{i}$, $i=1,...,c$, from the training set. In this case, the
trained weights or kernels of the Nadaraya-Watson regression are used to
estimate $\tilde{y}$. The number of subnetworks for testing is equal to the
number of training examples $n$ in every subset. Since the trained subnetworks
are identical and have the same weights (parameters), then their number can be
arbitrary. In fact, a single subnetwork can be used in practice, but its
output depends on the pair of vectors $\mathbf{x}$ and $\mathbf{x}_{i}$.

Quite the same network called the treatment network is constructed for the
treatment group. In contrast to the control network, it consists of $m$
subnetworks with inputs in the form of pairs $(\mathbf{z},\mathbf{z}_{i})$. In
the same way, $M$ subsets of size $m$ are randomly selected from the training
set of treatments without the example $(\mathbf{z}_{j},h_{j})$, $j=1,...,t$.
The treatment network is trained on $M\cdot t$ examples of size $2d\cdot m$.
After training, the treatment network allows us to estimate $\tilde{h}$ as
function $g_{1}(\mathbf{z})$. If $\mathbf{z}=\mathbf{x}$, then we get an
estimate of CATE $\tau(\mathbf{x})$ or $\tau(\mathbf{z})$ as the difference
between estimates $\tilde{h}$ and $\tilde{y}$ obtained by using the treatment
and control neural networks. It is important to point out that the control and
treatment networks are jointly trained by using the loss function defined below.

Let us formally describe TNW-CATE in detail. Consider the control group of
patients $\mathcal{C}=\{(\mathbf{x}_{1},y_{1}),...,(\mathbf{x}_{c},y_{c})\}$.
For every $i$ from set $\{1,...,c\}$, we define $N$ subsets $\mathcal{C}%
_{i,r}$, $r=1,...,N$, consisting of $n$ examples randomly selected from
$\mathcal{C}\backslash(\mathbf{x}_{i},y_{i})$ as:
\begin{equation}
\mathcal{C}_{i,r}=\{(\mathbf{x}_{1}^{(r)},y_{1}^{(r)}),...,(\mathbf{x}%
_{n}^{(r)},y_{n}^{(r)})\},\ r=1,...,N, \label{THTE_40}%
\end{equation}
where $\mathbf{x}_{j}^{(r)}$ is a randomly selected vector of features from
the set $\{\mathbf{x}_{1},...,\mathbf{x}_{c}\}\backslash\mathbf{x}_{i}$, which
forms $\mathcal{C}_{i,r}$; $y_{j}^{(r)}$ is the corresponding outcome.

Each subsets $\mathcal{C}_{i,r}$ jointly with $(\mathbf{x}_{i},y_{i})$ forms a
training example for the control network as follows:
\begin{equation}
\mathbf{a}_{i}^{(r)}=\left(  \mathbf{x}_{1}^{(r)},...,\mathbf{x}_{n}%
^{(r)},\mathbf{x}_{i},y_{1}^{(r)},...,y_{n}^{(r)},y_{i}\right)
,\ i=1,...,c,\ r=1,...,N. \label{THTE_42}%
\end{equation}

If we feed this example to the control network, then we expect to get some
approximation $\tilde{y}_{i}^{(r)}$ of $y_{i}$. The number of the above
examples for training is $N\cdot c$.

Let us consider the treatment group of patients $\mathcal{T}=\{(\mathbf{z}%
_{1},h_{1}),...,(\mathbf{z}_{t},h_{t})\}$ now. For every $j$ from set
$\{1,...,t\}$, we define $M$ subsets $\mathcal{T}_{j,s}$, $s=1,...,M$,
consisting of $m$ examples randomly selected from $\mathcal{T}\backslash
(\mathbf{z}_{j},h_{j})$ as:
\begin{equation}
\mathcal{T}_{j,s}=\{(\mathbf{z}_{1}^{(s)},h_{1}^{(s)}),...,(\mathbf{z}%
_{m}^{(s)},h_{m}^{(s)})\},\ s=1,...,M, \label{THTE_44}%
\end{equation}
where $\mathbf{z}_{l}^{(s)}$ is a randomly selected vector of features from
the set $\{\mathbf{z}_{1},...,\mathbf{z}_{t}\}\backslash\mathbf{z}_{j}$, which
forms $\mathcal{T}_{j,s}$; $h_{j}^{(s)}$ is the corresponding outcome.

Each subsets $\mathcal{T}_{j,s}$ jointly with $(\mathbf{z}_{j},h_{j})$ forms a
training example for the control network as follows:
\begin{equation}
\mathbf{b}_{j}^{(s)}=\left(  \mathbf{z}_{1}^{(s)},...,\mathbf{z}_{m}%
^{(s)},\mathbf{z}_{j},h_{1}^{(s)},...,h_{m}^{(s)},h_{j}\right)
,\ j=1,...,t,\ s=1,...,M. \label{THTE_46}%
\end{equation}

If we feed this example to the treatment network, then we expect to get some
approximation $\tilde{h}_{j}^{(s)}$ of $h_{j}$. Indices $r$ and $s$ are used
to distinguish subsets of controls and treatments.

\begin{figure}
[ptb]
\begin{center}
\includegraphics[
height=5.5679in,
width=4.8564in
]%
{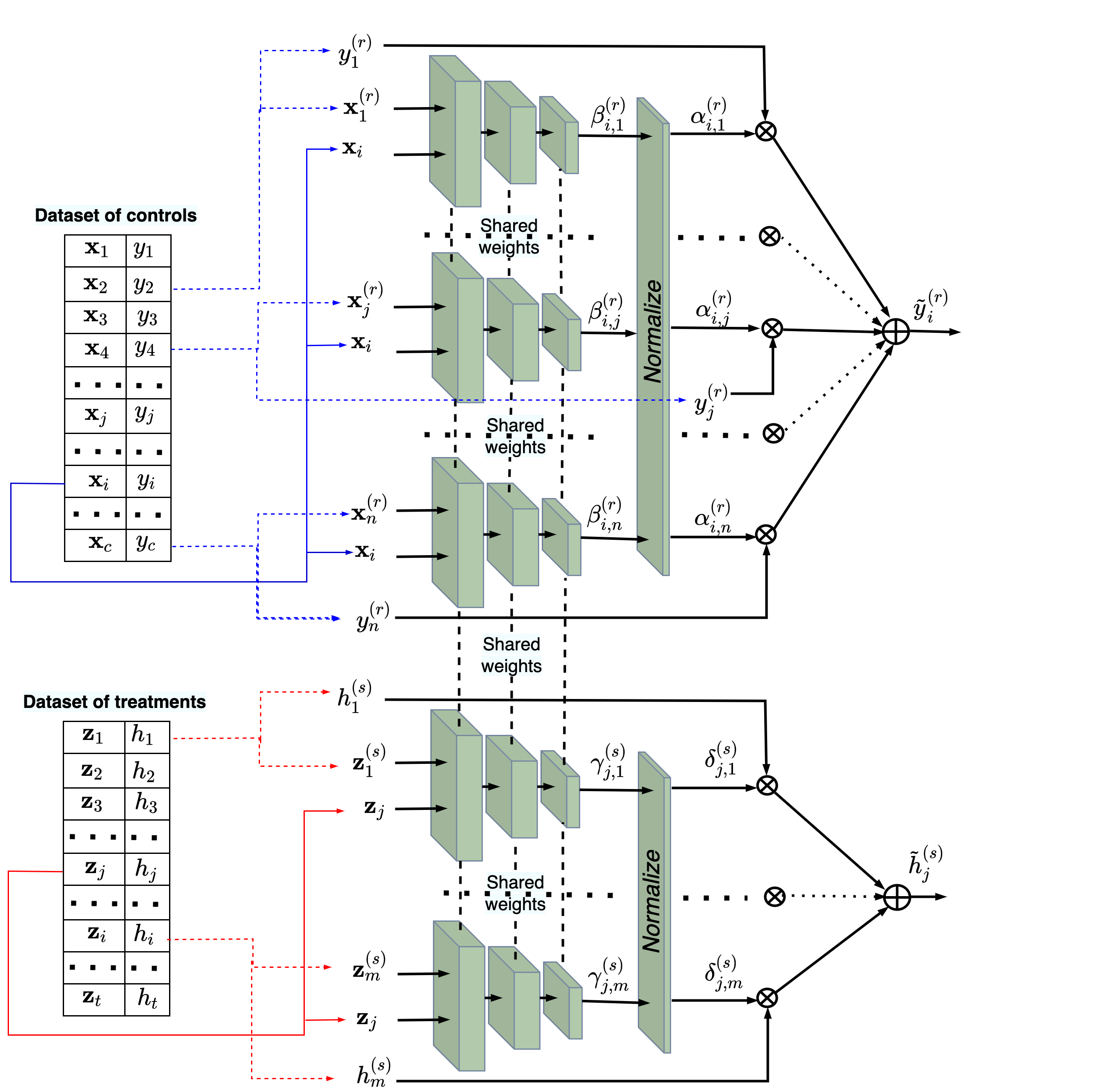}%
\caption{The weight sharing neural network consisting of $n$ subnetworks for
training on controls and $m$ subnetworks for training on treatments, which
implements two models of the Nadaraya-Watson regression}%
\label{f:train1}%
\end{center}
\end{figure}

The architecture of the joint neural network consisting of the control and
treatment networks is shown in Fig. \ref{f:train1}. One can see from Fig.
\ref{f:train1} that normalized outputs $\alpha_{i,j}^{(r)}$ and $\delta
_{j,l}^{(s)}$ of the subnetworks in the control and treatment networks are
multiplied by $y_{j}^{(r)}$ and $h_{l}^{(s)}$, respectively, and then the
obtained results are summed. Here $\alpha_{i,1}^{(r)}+...+\alpha_{i,n}%
^{(r)}=1$ and $\delta_{j,1}^{(s)}+...+\delta_{j,m}^{(s)}=1$. It should be
noted again that the control and treatment networks have the same parameters
(weights). Every network implements the Nadaraya-Watson regression with this
architecture, i.e.,
\begin{equation}
\tilde{y}_{i}^{(r)}=g_{0}(\mathbf{x}_{i})=\sum_{j=1}^{n}\alpha_{i,j}%
^{(r)}y_{j}^{(k)},\ \tilde{h}_{j}^{(s)}=g_{1}(\mathbf{z}_{j})=\sum_{j=1}%
^{n}\delta_{i,j}^{(s)}h_{j}^{(s)},
\end{equation}
Here
\begin{equation}
\alpha_{i,j}^{(r)}=\alpha(\mathbf{x}_{i},\mathbf{x}_{j}^{(r)})=\frac{K\left(
\mathbf{x}_{i},\mathbf{x}_{j}^{(r)}\right)  }{\sum_{k=1}^{n}K\left(
\mathbf{x}_{i},\mathbf{x}_{k}^{(r)}\right)  },
\end{equation}%
\begin{equation}
\delta_{i,j}^{(s)}=\delta(\mathbf{z}_{j},\mathbf{z}_{i}^{(s)})=\frac{K\left(
\mathbf{z}_{j},\mathbf{z}_{i}^{(s)}\right)  }{\sum_{k=1}^{m}K\left(
\mathbf{z}_{j},\mathbf{z}_{k}^{(s)}\right)  }.
\end{equation}

If to consider the whole neural network, then training examples are of the
form:
\begin{equation}
\left(  \mathbf{a}_{i}^{(r)},\mathbf{b}_{j}^{(s)}\right)
,\ i=1,...,c,\ j=1,...,t,~r=1,...,N,\ s=1,...,M.
\end{equation}

The standard expected $L_{2}$ loss function for training the whole network is
of the form:
\begin{align}
L  &  =\frac{1}{N\cdot c}\sum_{r=1}^{N}\sum_{i=1}^{c}\left(  \tilde{y}%
_{i}^{(r)}-y_{i}\right)  ^{2}+\alpha\frac{1}{M\cdot t}\sum_{s=1}^{M}\sum
_{j=1}^{t}\left(  \tilde{h}_{j}^{(s)}-h_{j}\right)  ^{2}\nonumber\\
&  =\frac{1}{N\cdot c}\sum_{r=1}^{N}\sum_{i=1}^{c}\left(  g_{0}\left(
\mathbf{a}_{i}^{(r)}\right)  -y_{i}\right)  ^{2}+\alpha\frac{1}{M\cdot t}%
\sum_{s=1}^{M}\sum_{j=1}^{t}\left(  g_{1}\left(  \mathbf{b}_{j}^{(s)}\right)
-h_{j}\right)  ^{2}. \label{THTE_70}%
\end{align}

Here $\alpha$ is the coefficient which controls how the treatment networks
impacts on the training process. In particular, if $\alpha=0$, then only the
control network is trained on controls without treatments.

\begin{figure}
[ptb]
\begin{center}
\includegraphics[
height=2.8222in,
width=4.4738in
]%
{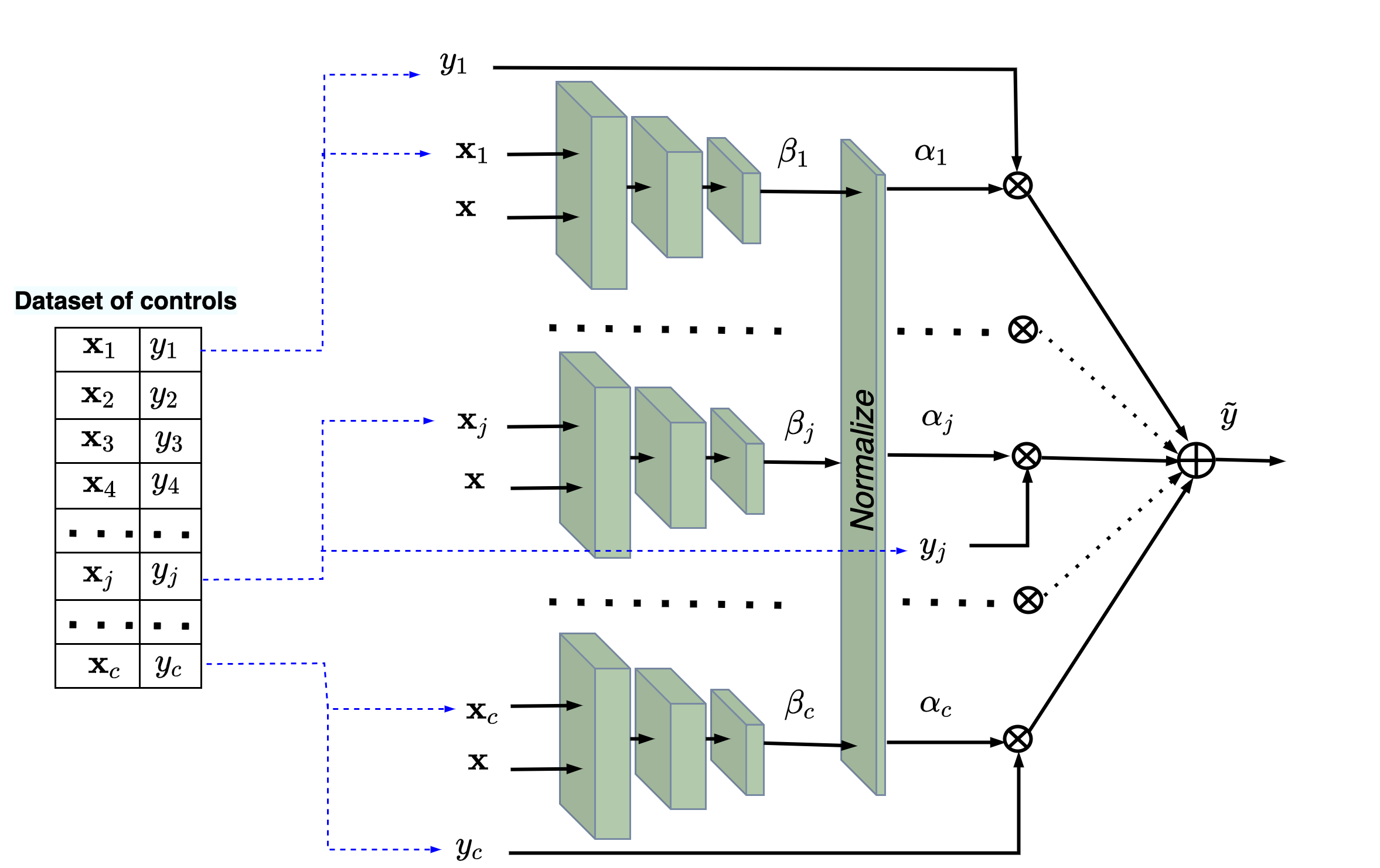}%
\caption{The trained neural network for computing $\tilde{y}$ in accordance
with the Nadaraya-Watson regression}%
\label{f:test_contr}%
\end{center}
\end{figure}

\begin{figure}
[ptb]
\begin{center}
\includegraphics[
height=2.2127in,
width=4.4721in
]%
{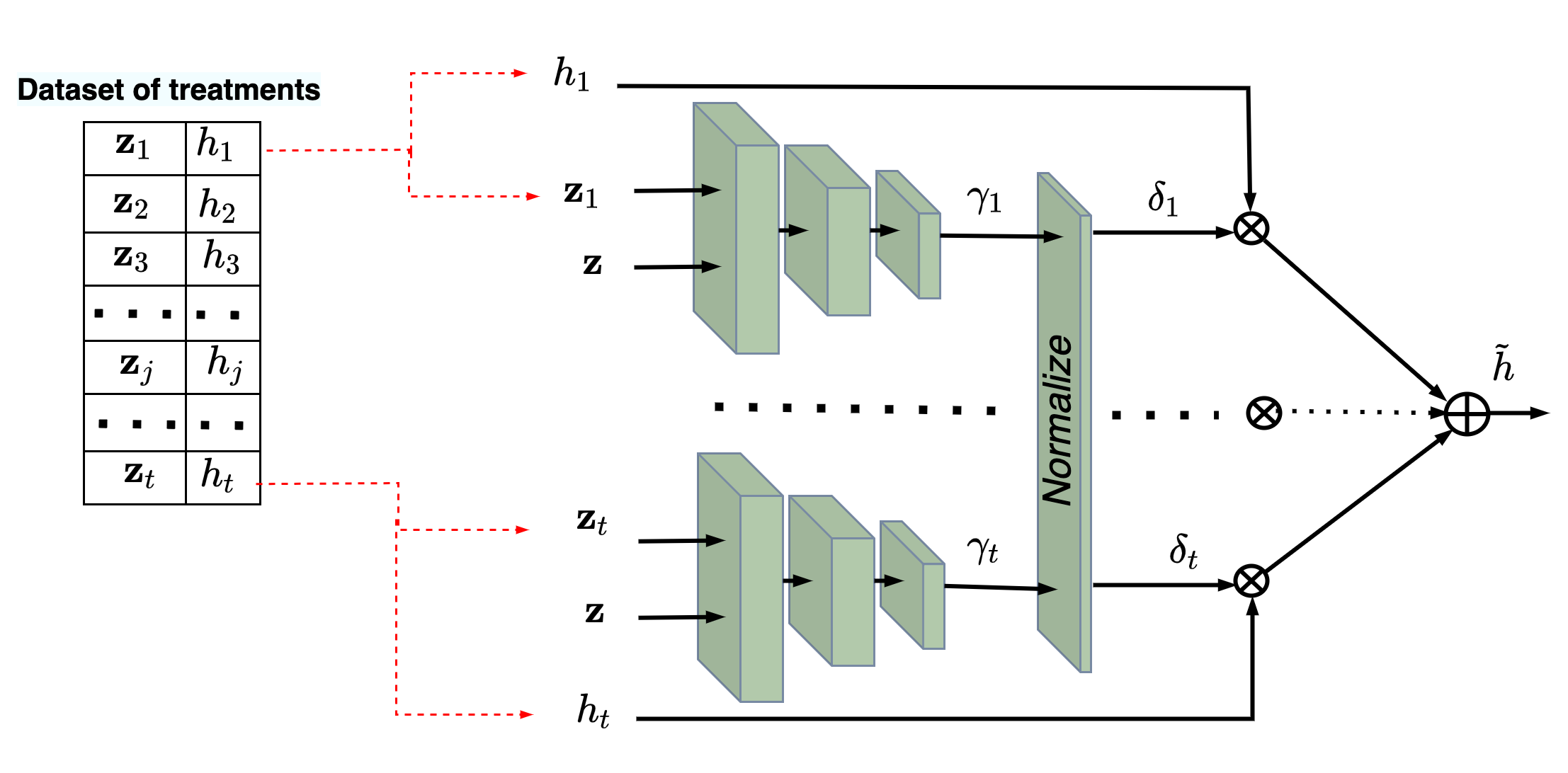}%
\caption{The trained neural network for computing $\tilde{h}$ in accordance
with the Nadaraya-Watson regression}%
\label{f:test_treat}%
\end{center}
\end{figure}

In sum, we achieve our first goal to train subnetworks implementing kernels in
the Nadaraya-Watson regression by using examples from control and treatment
groups. The trained kernels take into account structures of the treatment and
control data. The next task is to estimate CATE by using the trained kernels
for some new vectors $\mathbf{x}$ or $\mathbf{z}$. It should be noted that all
subnetworks can be represented as a single network due to shared weights. In
this case, arbitrary batches of pairs $(\mathbf{x},\mathbf{x}_{i})$ and pairs
$(\mathbf{z},\mathbf{z}_{j})$ can be fed to the single network. This implies
that we can construct testing neural networks consisting of $c$ and $t$
trained subnetworks in order to estimate $\tilde{y}$ and $\tilde{h}$
corresponding to $\mathbf{z}$ and $\mathbf{x}$, respectively, under condition
$\mathbf{z}=\mathbf{x}$. Figs. \ref{f:test_contr} and \ref{f:test_treat} show
trained neural networks for estimating $\tilde{y}$ and $\tilde{h}$,
respectively. It is important to point out that the testing networks are not
trained. Sets of $c$ pairs $(\mathbf{x},\mathbf{x}_{1}),...,(\mathbf{x}%
,\mathbf{x}_{c})$ and $t$ pairs $(\mathbf{z},\mathbf{z}_{1}),...,(\mathbf{z}%
,\mathbf{z}_{t})$ are fed to the subnetworks of the control and treatment
networks, respectively. The whole examples for testing taking into account
outcomes are
\begin{equation}
\mathbf{a}(\mathbf{x})=\left(  \mathbf{x}_{1},...,\mathbf{x}_{c}%
,\mathbf{x},y_{1},...,y_{c}\right)  , \label{THTE_74}%
\end{equation}
and
\begin{equation}
\mathbf{b}(\mathbf{z})=\left(  \mathbf{z}_{1},...,\mathbf{z}_{c}%
,\mathbf{z},h_{1},...,h_{t}\right)  . \label{THTE_75}%
\end{equation}

In sum, the networks implement the Nadaraya-Watson regressions:
\begin{equation}
\tilde{y}=g_{0}(\mathbf{x})=\sum_{j=1}^{c}\alpha(\mathbf{x},\mathbf{x}%
_{j})y_{j},\ \tilde{h}=g_{1}(\mathbf{z})=\sum_{i=1}^{t}\delta(\mathbf{z}%
,\mathbf{z}_{i})h_{i}.
\end{equation}

Finally, CATE $\tau(\mathbf{x})$ or $\tau(\mathbf{z})$ is estimated as
$\tau(\mathbf{x})=\tau(\mathbf{z})=\tilde{h}-\tilde{y}$.%

%

%

The phases of the neural network training and testing are schematically shown
as Algorithms \ref{alg:Univers_Train} and \ref{alg:Univers_Test}, respectively.

\begin{algorithm}
\caption{The algorithm implementing TNW-CATE in the training phase}
\label{alg:Univers_Train}

\begin{algorithmic}
[1]\REQUIRE Datasets $\mathcal{C}$ of $c$ controls and $\mathcal{T}$ of $t$
treatments, numbers $N$ and $M$ of generated subsets of $\mathcal{C}$ and
$\mathcal{T}$, numbers of examples in generated subsets $n$ and $m$

\ENSURE The trained weight sharing neural network implementing the
Nadaraya-Watson regressions for control and treatment data

\FOR{$i=1$, $i\leq c$ }\FOR{$r=1$, $r\leq N$ }

\STATE Generate subset $\mathcal{C}_{i,r}\subset\mathcal{C}\backslash
(\mathbf{x}_{i},y_{i})$ (see (\ref{THTE_40}))

\STATE Form example $\mathbf{a}_{i}^{(r)}$ (see (\ref{THTE_42}))

\ENDFOR\ENDFOR

\FOR{$j=1$, $j\leq t$ }\FOR{$s=1$, $s\leq M$ }

\STATE Generate subset $\mathcal{T}_{j,s}\subset\mathcal{T}\backslash
(\mathbf{z}_{j},h_{j})$ (see (\ref{THTE_44}))

\STATE Form example $\mathbf{b}_{j}^{(s)}$ (see (\ref{THTE_46}))

\ENDFOR\ENDFOR

\STATE Train the weight sharing neural network with the loss function given in
(\ref{THTE_70}) on the set of pairs $(\mathbf{a}_{i}^{(r)},\mathbf{b}%
_{j}^{(s)})$
\end{algorithmic}
\end{algorithm}

\begin{algorithm}
\caption{The algorithm implementing TNW-CATE in the testing phase}
\label{alg:Univers_Test}

\begin{algorithmic}
[1]\REQUIRE Trained neural control and treatment networks implementing the
Nadaraya-Watson regressions; datasets $\mathcal{C}$ and $\mathcal{T}$; testing
example $\mathbf{x}=\mathbf{z}$

\ENSURE CATE $\tau(\mathbf{x})$

\STATE Form the testing example $\mathbf{a}(\mathbf{x})$ in accordance with
(\ref{THTE_74})

\STATE Form the testing example $\mathbf{b}(\mathbf{z})$ in accordance with
(\ref{THTE_75})

\STATE Feed $\mathbf{a}(\mathbf{x})$ to the control network and $\mathbf{b}%
(\mathbf{z})$ to the treatment network and get the corresponding predictions
$\tilde{y}$ and $\tilde{h}$

\STATE$\tau(\mathbf{x})=\tilde{h}-\tilde{y}$
\end{algorithmic}
\end{algorithm}

It is important to point out that the neural networks shown in Figs.
\ref{f:test_contr} and \ref{f:test_treat} are not trained on datasets
$\mathcal{C}$ and $\mathcal{T}$. These datasets are used as testing examples.
This is an important difference of the proposed approach from other
classification or regression models.

Let us return to the case $\alpha=0$ when only the control network is trained
on controls without treatments. This case is interesting because it clearly
demonstrates the transfer learning model when domains of source and target
data are the same, but tasks are different. Indeed, we train kernels of the
Nadaraya-Watson regression on controls under assumption that domains of
controls and treatments are the same. Kernels learn the feature vector
location. Actually, kernels are trained on controls by using outcomes $y_{i}$.
However, nothing prevents us from using the same kernels with different
outcomes $h_{i}$ if structures of feature vectors in the control and treatment
groups are similar. We often use the same standard kernels with the same
parameters, for example, Gaussian ones with the temperature parameter, in
machine learning tasks. The proposed method does the same, but with a more
complex kernel.

\section{Numerical experiments}

In this section, we provide simulation experiments evaluating the performance
of meta-models for CATE estimation. In particular, we compare the T-learner,
the S-learner, the X-learner and the proposed method in several simulation
studies. Numerical experiments are based on random generation of control and
treatment data in accordance with different predefined control and treatment
outcome functions because the true CATEs are unknown due to the fundamental
problem of the causal inference for real data \cite{Kunzel-etal-2018a}.

\subsection{General parameters of experiments}

\subsubsection{CATE estimators for comparison}

The following models are used for their comparison with TNW-CATE.

\begin{enumerate}
\item The T-learner \cite{Kunzel-etal-2018} is a simple procedure based on
estimating the control $g_{0}(\mathbf{x})$ and treatment $g_{1}(\mathbf{z})$
outcome functions by applying a regression algorithm, for example, the random
forest \cite{Breiman-2001}. The CATE in this case is defined as the difference
$g_{1}(\mathbf{z})-g_{0}(\mathbf{x})$.

\item The S-learner was proposed in \cite{Kunzel-etal-2018} to overcome
difficulties and disadvantages of the T-learner. The treatment assignment
indicator $T_{i}$ in the S-learner is included as an additional feature to the
feature vector $\mathbf{x}$. The corresponding training set in this case is
modified as $\mathcal{D}^{\ast}=\{(\mathbf{x}_{1}^{\ast},y_{1}%
),...,(\mathbf{x}_{c+t}^{\ast},y_{c+t})\}$, where $\mathbf{x}_{i}^{\ast
}=(\mathbf{x}_{i},T_{i})\in\mathbb{R}^{d+1}$ if $T_{i}=0$, $i=1,...,c$, and
$\mathbf{x}_{c+i}^{\ast}=(\mathbf{z}_{i},T_{i})\in\mathbb{R}^{d+1}$ if
$T_{i}=1$, $i=1,...,t$. Then the outcome function $g(\mathbf{x},T)$ is
estimated by using the training set $\mathcal{D}^{\ast}$. The CATE is
determined in this case as $\tau(\mathbf{x})=g(\mathbf{x},1)-g(\mathbf{x},0).$

\item The X-learner \cite{Kunzel-etal-2018} is based on computing the
so-called imputed treatment effects and is represented in the following three
steps. First, the outcome functions $g_{0}(\mathbf{x})$ and $g_{1}%
(\mathbf{z})$ are estimated using a regression algorithm, for example, the
random forest. Second, the imputed treatment effects are computed as follows:
\begin{equation}
D_{1}(\mathbf{z}_{i})=h_{i}-g_{0}(\mathbf{z}_{i}),\ \ D_{0}(\mathbf{x}%
_{i})=g_{1}(\mathbf{x}_{i})-y_{i}.
\end{equation}

Third, two regression functions $\tau_{1}(\mathbf{z})$ and $\tau
_{0}(\mathbf{x})$ are estimated for imputed treatment effects $D_{1}%
(\mathbf{z}_{i})$ and $D_{0}(\mathbf{x}_{i})$, respectively. CATE for a point
$\mathbf{x}=\mathbf{z}$\ is defined as a weighted linear combination of the
functions $\tau_{1}(\mathbf{z})$ and $\tau_{0}(\mathbf{x})$ as $\tau
(\mathbf{x})=\alpha\tau_{0}(\mathbf{x})+(1-\alpha)\tau_{1}(\mathbf{x})$, where
$\alpha\in\lbrack0,1]$ is a weight which is equal to the ratio of treated
patients \cite{Kunzel-etal-2018a}.
\end{enumerate}

\subsubsection{Base models for implementing estimators}

Two models are used as the base regressors $g_{0}(\mathbf{x})$ and
$g_{1}(\mathbf{z})$, which realize different CATE estimators for comparison purposes.

\begin{enumerate}
\item The first one is the random forest. It is used as the base regressor to
implement other models due to two main reasons. First, we consider the case of
the small number of treatments, and usage of neural networks does not allow us
to obtain the desirable accuracy of the corresponding regressors. Second, we
deal with tabular data for which it is difficult to train a neural network and
the random forest is preferable. Parameters of random forests used in
experiments are the following:

\begin{itemize}
\item numbers of trees are 10, 50, 100, 300;

\item depths are 2, 3, 4, 5, 6, 7;

\item the smallest values of examples which fall in a leaf are 1 example, 5\%,
10\%, 20\% of the training set.
\end{itemize}

The above values for the hyperparameters are tested, choosing those leading to
the best results.

\item The second base model used for realization different CATE estimators is
the Nadaraya-Watson regression with the standard Gaussian kernel. This model
is used because it is interesting to compare it with the proposed model which
is also based on the Nadaraya-Watson regression but with the trainable kernel
in the form of the neural network of the special form. Values $10^{i}$,
$i=-8,...,10$, and also values $0.5$, $5$, $50$, $100$, $200$, $500$, $700$ of
the bandwidth parameter $\gamma$ are tested, choosing those leading to the
best results.
\end{enumerate}

We use the following notation for different models depending on the base
models and learners:

\begin{itemize}
\item \textbf{T-RF, S-RF, X-RF} are the T-learner, the S-learner, the
X-learner with random forests as the base regression models;

\item \textbf{T-NW, S-NW, X-NW} are the T-learner, the S-learner, the
X-learner with the Nadaraya-Watson regression using the standard Gaussian
kernel as the base regression models.
\end{itemize}

\subsubsection{Other parameters of numerical experiments}

The mean squared error (MSE) as a measure of the regression model accuracy is
used. For estimating MSE, we perform several iterations of experiments such
that each iteration is defined by the randomly selected parameters of
experiments. MSE is computed by using 1000 points. In all experiments, the
number of treatments is $10\%$ of the number of controls. For example, if
$100$ controls are generated for an experiment, then $10$ treatments are
generated in addition to controls such that the total number of examples is
$110$. After generating training examples, their responses $y$ are normalized,
but the corresponding initial mean and the standard deviation of responses are
used to normalize responses of the test examples. This procedure allows us to
reduce the variance among results at different iterations. The generated
feature vectors in all experiments consist of $10$ features. To select optimal
hyperparameters of all regressors, additional validation examples are
generated such that the number of controls is $20\%$ of the training examples
from the control group. The validation examples are not used for training.

\subsubsection{Functions for generating datasets}

The following functions are used to generate synthetic datasets:

\begin{enumerate}
\item \textbf{Spiral functions:} The functions are named spiral because for
the case of two features vectors are located on the Archimedean spiral. For
even $d$, we write the vector of features as
\begin{equation}
\mathbf{x}=(t\sin(t),t\cos(t),...,t\sin(t\cdot d/2),t\cos(t\cdot d/2)).
\end{equation}

For odd $d$, there holds
\begin{equation}
\mathbf{x}=(t\sin(t),t\cos(t),...,t\sin(t\cdot\left\lceil d/2\right\rceil )).
\end{equation}

The responses are generated as a linear function of $t$, i.e., they are
computed as $y=at+b$.

Values of parameters $a$, $b$ and $t$ for performing numerical experiments
with spiral functions are the following:

\begin{itemize}
\item The control group: parameters $a$, $b$, $t$ are uniformly generated from
intervals $[1,4]$, $[1,4]$, $[0,10]$, respectively.

\item The treatment group: parameters $a$, $b$, $t$ are uniformly generated
from intervals $[8,10]$, $[8,10]$, $[0,10]$, respectively.
\end{itemize}

\item \textbf{Logarithmic functions:} Features are logarithms of the parameter
$t$, i.e., there holds
\begin{equation}
\mathbf{x}=(a_{1}\ln(t),a_{2}\ln(t),...,a_{d}\ln(t)).
\end{equation}
The responses are generated as a logarithmic function with adding an
oscillating term $b\sin(t)$ to $y$, i.e., there holds $y=b(\ln(t)+\sin(t))$.

Values of parameters $a_{1},...,a_{d}$, $b$ for performing numerical
experiments with logarithmic functions are the following:

\begin{itemize}
\item Each parameter from the set $\{a_{1},...,a_{d}\}$ is uniformly generated
from intervals $[-4,-1]\cup\lbrack1,4]$ for controls as well as for treatments.

\item Parameter $b$ is uniformly generated from interval $[1,4]$ for controls
and from interval $[-4,-1]$ for treatments.

\item Values of $t$ are uniformly generated in interval $[0.5,5]$.
\end{itemize}

\item \textbf{Power functions:} Features are represented as powers of $t$. For
arbitrary $d$, the vector of features is represented as
\begin{equation}
\mathbf{x}=(t^{1/\sqrt{d}},t^{2/\sqrt{d}},...,t^{d/\sqrt{d}}).
\end{equation}
However, features which are close to linear ones, i.e., $x_{i}=t^{i/\sqrt{d}}$
for $0.8<i/\sqrt{d}<1.6$, are replaced with the Gaussian noise having the unit
standard deviation and the zero expectation, i.e., $x_{i}\sim\mathcal{N}%
(0,1)$. The responses are generated as follows:%
\begin{equation}
y=a\cdot\exp\left(  -\frac{(t-s)^{2}}{b}\right)  .
\end{equation}

Values of parameters $a$, $b$, $s$, $t$ for performing numerical experiments
with power functions are the following:

\begin{itemize}
\item The control group: parameters $a$ and $b$ are uniformly generated from
intervals $[1,2]$ and $[0.25,1]$, respectively; parameter $s$ is $2.5$.

\item The treatment group: parameters $a$ and $b$ are uniformly generated from
intervals $[2,4]$ and $[1,2]$, respectively; parameter $s$ is $2.5$.

\item Values of $t$ are uniformly generated in interval $[0,5]$.
\end{itemize}

\item \textbf{Indicator functions} \cite{Kunzel-etal-2018}: The functions are
expressed through the indicator function $I$ taking value $1$ if its argument
is true.

\begin{itemize}
\item The function for controls is represented as
\begin{equation}
g_{0}(\mathbf{x})=\mathbf{x}^{\mathrm{T}}\beta+5I(x_{1}>0.5). \label{THTE_85}%
\end{equation}

\item The function for treatments is represented as
\begin{equation}
g_{1}(\mathbf{x})=\mathbf{x}^{\mathrm{T}}\beta+5I(x_{1}>0.5)+8I(x_{2}>0.1).
\label{THTE_86}%
\end{equation}

\item Vector $\beta$ is uniformly distributed in interval $[-5;5]^{d}$; values
of features $x_{i}=1,...,d$, are uniformly generated from interval $[-1,1]$.
\end{itemize}
\end{enumerate}

The indicator function differs from other functions considered in numerical
examples. It is taken from \cite{Kunzel-etal-2018} to study TNW-CATE when the
assumption of specific and similar domains for the control and treatment
feature vectors can be violated.

In numerical experiments with the above functions, we take parameter $\alpha$
equal to $0.1$, $0.5$, $0.5$ and $0.5$ for the spiral, logarithmic, power and
indicator functions, respectively, except for experiments which study how
parameter $\alpha$ impacts on the MSE.

\subsection{Study of the TNW-CATE properties}

In all pictures illustrating dependencies of CATE estimators on parameters of
models, dotted curves correspond to the T-learner, the S-learner, the
X-learner implemented by using the Nadaraya-Watson regression (triangle
markers correspond to T-NW and S-NW, the circle marker corresponds to X-NW).
Dashed curves with the same markers correspond to the same models implemented
by using RFs. The solid curve with cross markers corresponds to TNW-CATE.

\subsubsection{Experiments with numbers of training data}

\begin{figure}
[ptb]
\begin{center}
\includegraphics[
height=4.3279in,
width=5.3155in
]%
{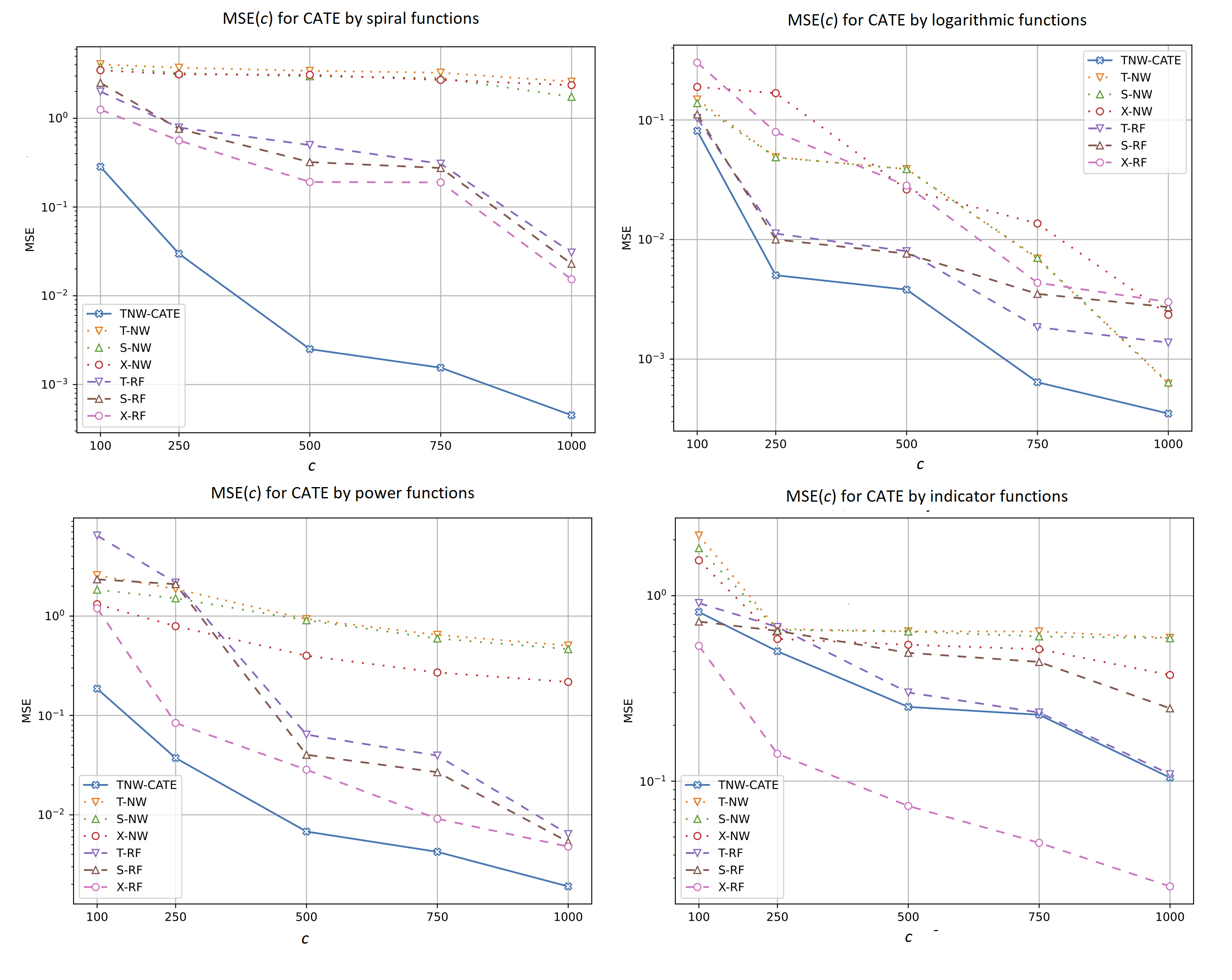}%
\caption{MSE of the CATE values as a function of the number of controls when
spiral, logarithmic, power and indicator functions are used for generating
examples}%
\label{f:size_cate}%
\end{center}
\end{figure}

Let us compare different CATE estimators by different numbers of control and
treatment examples. We study the estimators by numbers $c$ of controls: $100$,
$250$, $500$, $750$, $1000$. Each number of treatments is determined as $10\%$
of each number of controls. Value $n$ is $80$ and $100$, value $m$ is $50\%$
of $t$. Fig. \ref{f:size_cate} illustrates how MSE of the CATE values depends
on the number $c$ of controls for different estimators when different
functions are used for generating examples. In fact, these experiments study
how MSE depends on the whole number of controls and treatments because the
number of treatments increases with the number of controls. For all functions,
the increase of the amount of training examples improves most estimators
including TNW-CATE. These results are expectable because the larger size of
training data mainly leads to the better accuracy of models. It can be seen
from Fig. \ref{f:size_cate} that the proposed model provides better results in
comparison with other models. The best results are achieved when the spiral
generating function is used. The models different from TNW-CATE cannot cope
with the complex structure of data in this case. However, TNW-CATE shows
comparative results with the T-learner, the S-learner, the X-learner when the
indicator function is used for generating examples. The X-learner outperforms
TNW-CATE in this case. The indicator function does not have a complex
structure. Moreover, the corresponding outcomes linearly depend on most
features (see (\ref{THTE_85})-(\ref{THTE_86})) and random forests implementing
X-RF are trained better than the neural network implementing TNW-CATE. One can
also see from Fig. \ref{f:size_cate} that models T-NW, S-NW, X-NW based on the
Nadaraya-Watson regression with the Gaussian kernel provide close results when
the logarithmic generating function is used by larger numbers of training
data. This is caused by the fact that the Gaussian kernel is close to the
neural network kernel implemented in TNW-CATE.

\begin{figure}
[ptb]
\begin{center}
\includegraphics[
height=2.2303in,
width=5.4096in
]%
{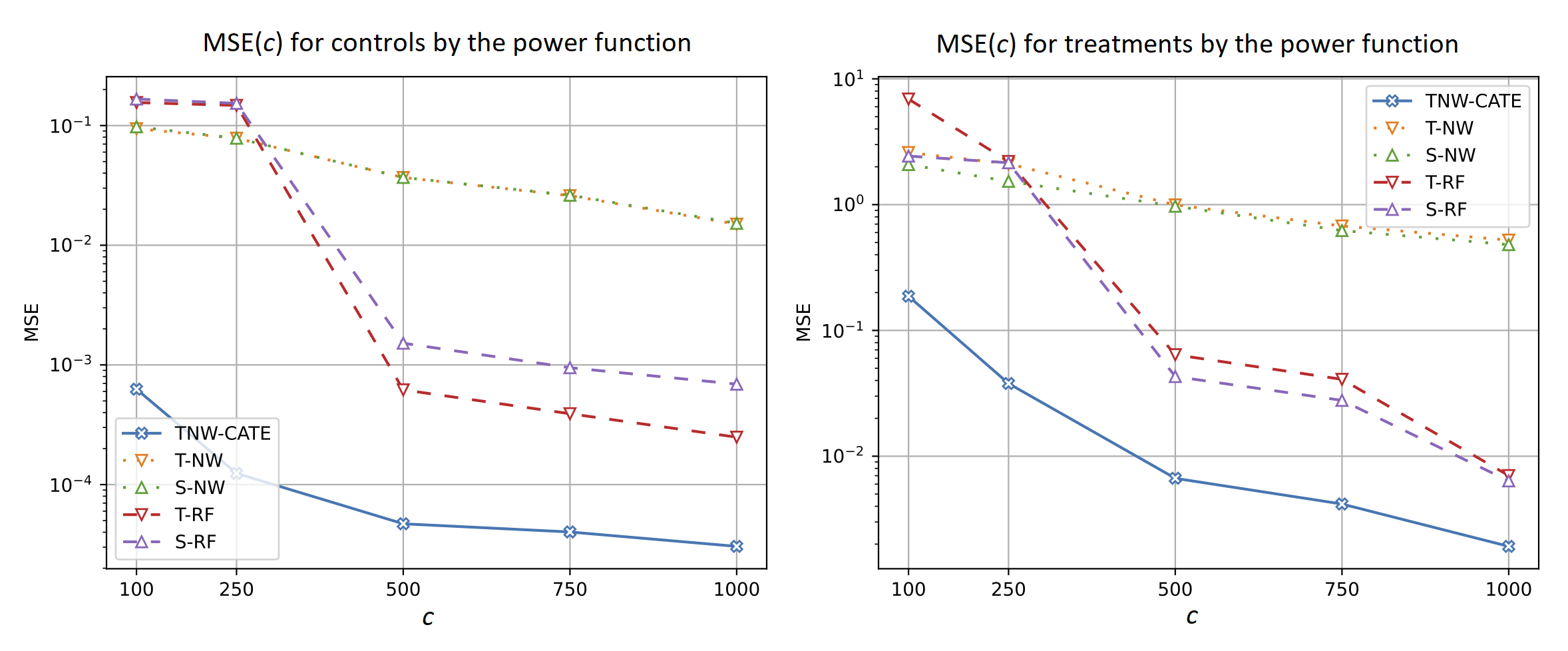}%
\caption{MSE of the control (the left plot) and treatment (the right plot)
responses as functions of the number of controls when the power function is
used for generating examples}%
\label{f:size_cont_treat_pow}%
\end{center}
\end{figure}

Fig. \ref{f:size_cont_treat_pow} illustrates how the number of controls
impacts separately on the control (the left plot) and treatment (the right
plot) regressions when the power function is used for generating data. It can
be also seen from Fig. \ref{f:size_cont_treat_pow} that MSE provided by the
control network is much smaller than MSE of the treatment neural network.%

%

\subsubsection{Experiments with different values of the treatment ratio}

Another question is how the CATE estimators depend on the ratio of numbers of
treatments and controls in the training set. We study the case when the number
of controls $c$ is $200$ and the ratio takes values from the set
$\{0.1,0.2,0.3,0.4,0.5\}$. The coefficient $\alpha$ is taken in accordance
with the certain function as described above.

\begin{figure}
[ptb]
\begin{center}
\includegraphics[
height=4.8678in,
width=5.5019in
]%
{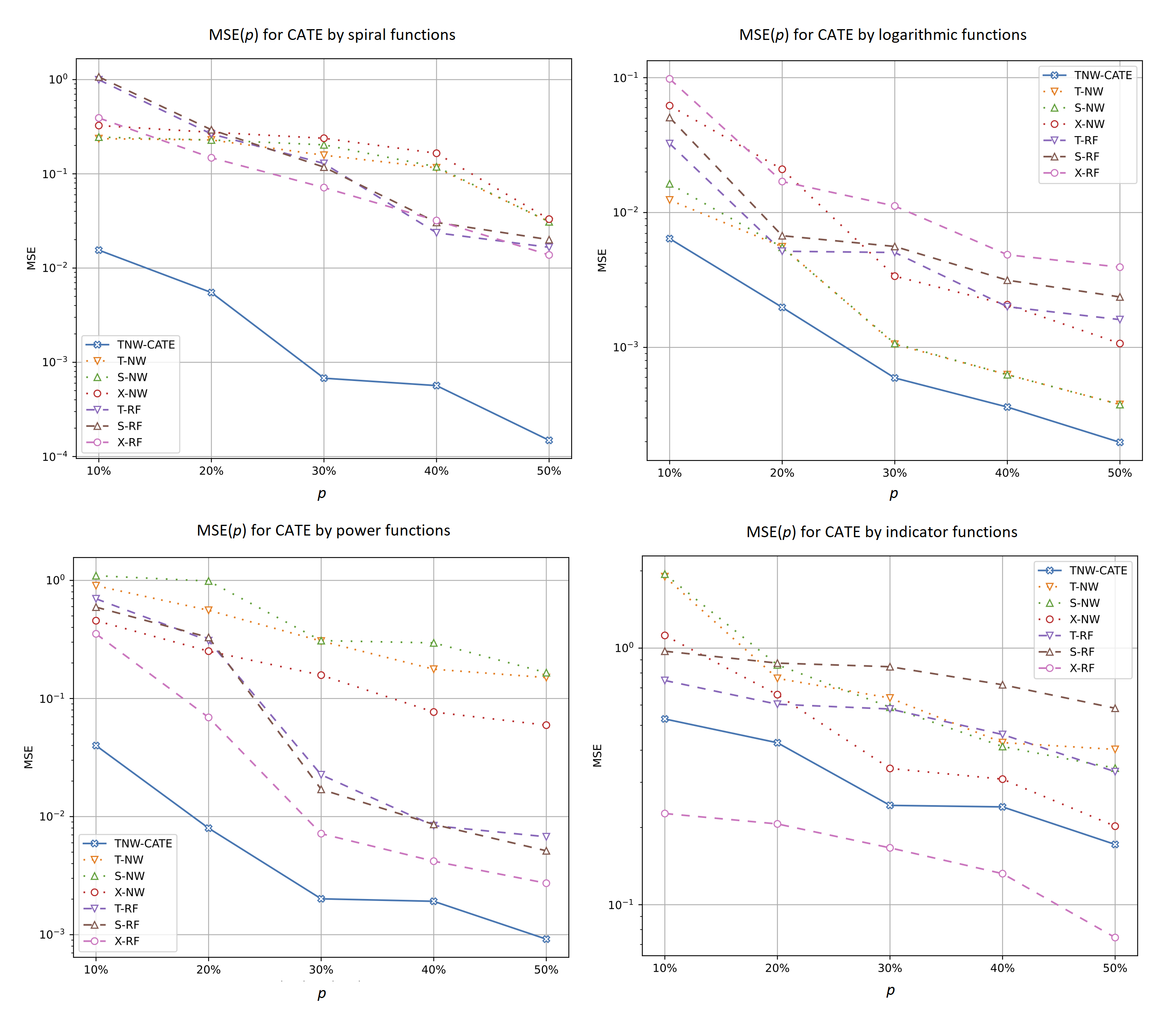}%
\caption{MSE of the CATE values as a function of the ratio of numbers of
treatments in the training set when spiral, logarithmic, power and indicator
functions are used for generating examples }%
\label{f:part_cate}%
\end{center}
\end{figure}

Similar results are shown in Fig. \ref{f:part_cate} where plots of MSE of the
CATE values as a function of the ratio of numbers of treatments in the
training set by different generating functions are depicted. We again see from
Fig. \ref{f:part_cate} that the difference between MSE of TNW-CATE and other
models is largest when the spiral function is used. TNW-CATE also provides
better results in comparison with other models, except for the case of the
indicator function when is TNW-CATE inferior to the X-RF.

\begin{figure}
[ptb]
\begin{center}
\includegraphics[
height=2.4159in,
width=5.8871in
]%
{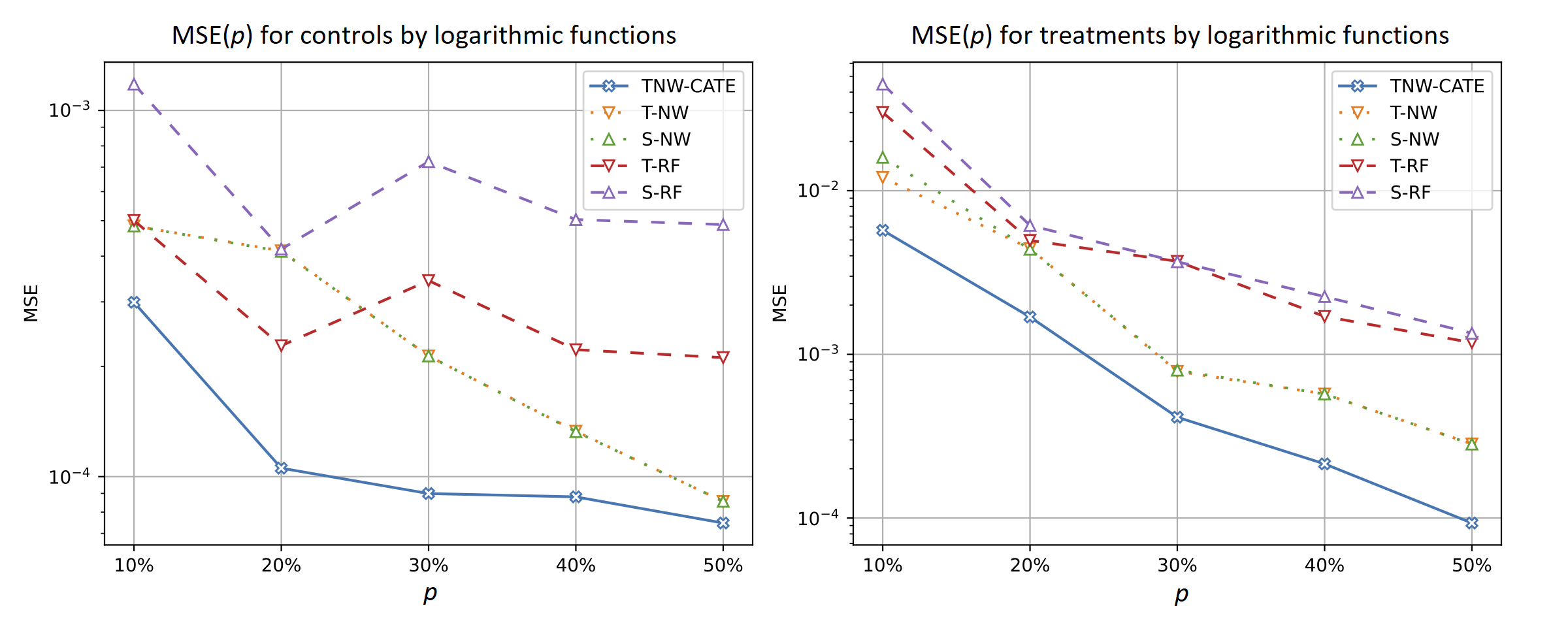}%
\caption{MSE of the control (the left plot) and treatment (the right plot)
responses as functions of the ratio of numbers of treatments in the training
set when the logarithmic function is used for generating examples}%
\label{f:part_cont_treat_log}%
\end{center}
\end{figure}

Fig. \ref{f:part_cont_treat_log} illustrates how the ratio of numbers of
treatments impacts separately on the control (the left plot) and treatment
(the right plot) regressions when the logarithmic function is used for
generating data. It can be seen from Fig. \ref{f:part_cont_treat_log} that MSE
of the treatment network is very close to MSE\ of other models almost for all
values of the ratio, but the accuracy of the control network significantly
differs from other model.%

%

\subsubsection{Experiments with different values of $\alpha$}

\begin{figure}
[ptb]
\begin{center}
\includegraphics[
height=5.4052in,
width=4.684in
]%
{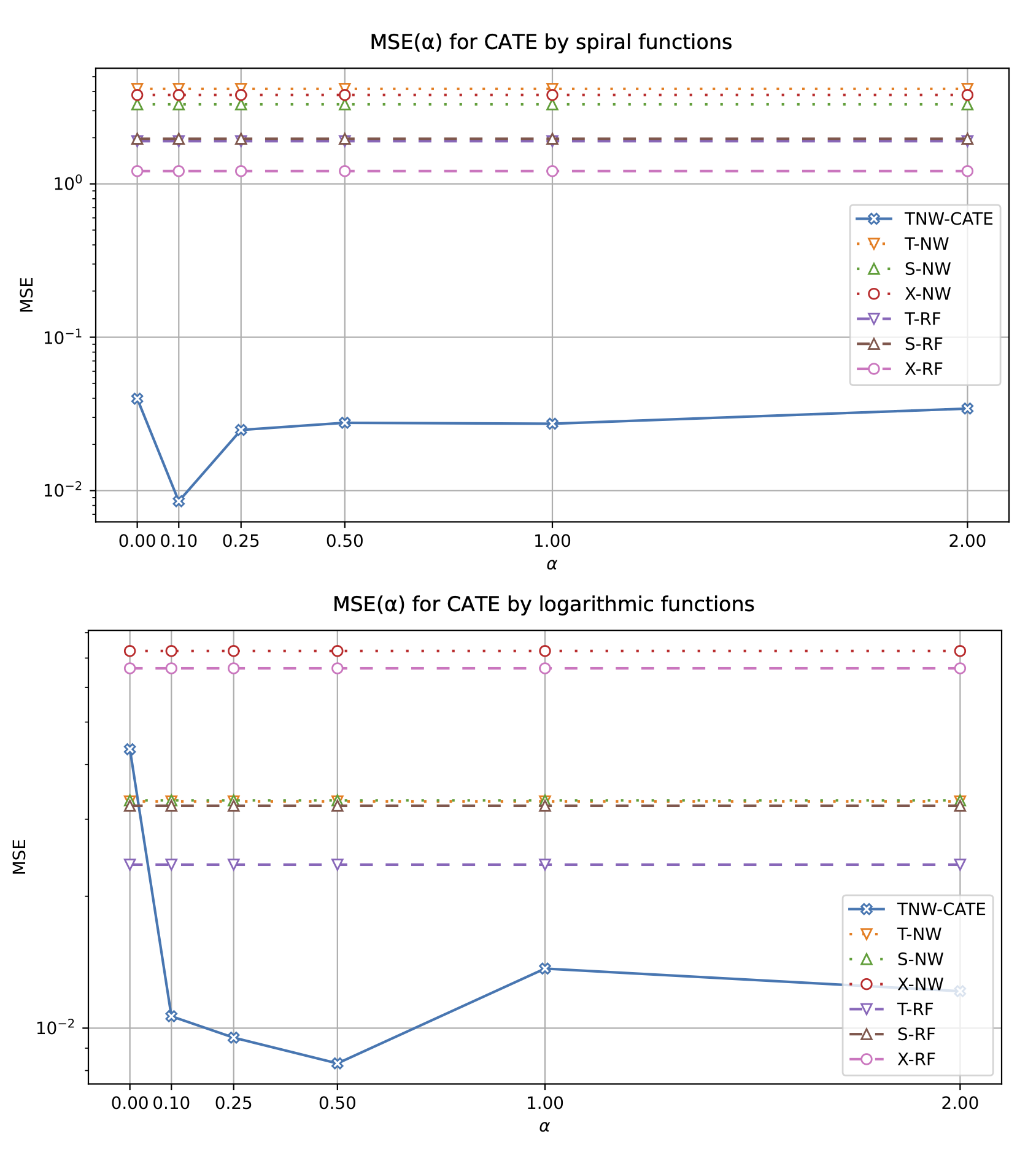}%
\caption{MSE of the CATE values as a function of the coefficient $\alpha$ when
spiral (the first plot) and logarithmic (the second plot) functions are used
for generating examples}%
\label{f:alpha_cate_srl_log}%
\end{center}
\end{figure}

\begin{figure}
[ptb]
\begin{center}
\includegraphics[
height=5.246in,
width=4.7509in
]%
{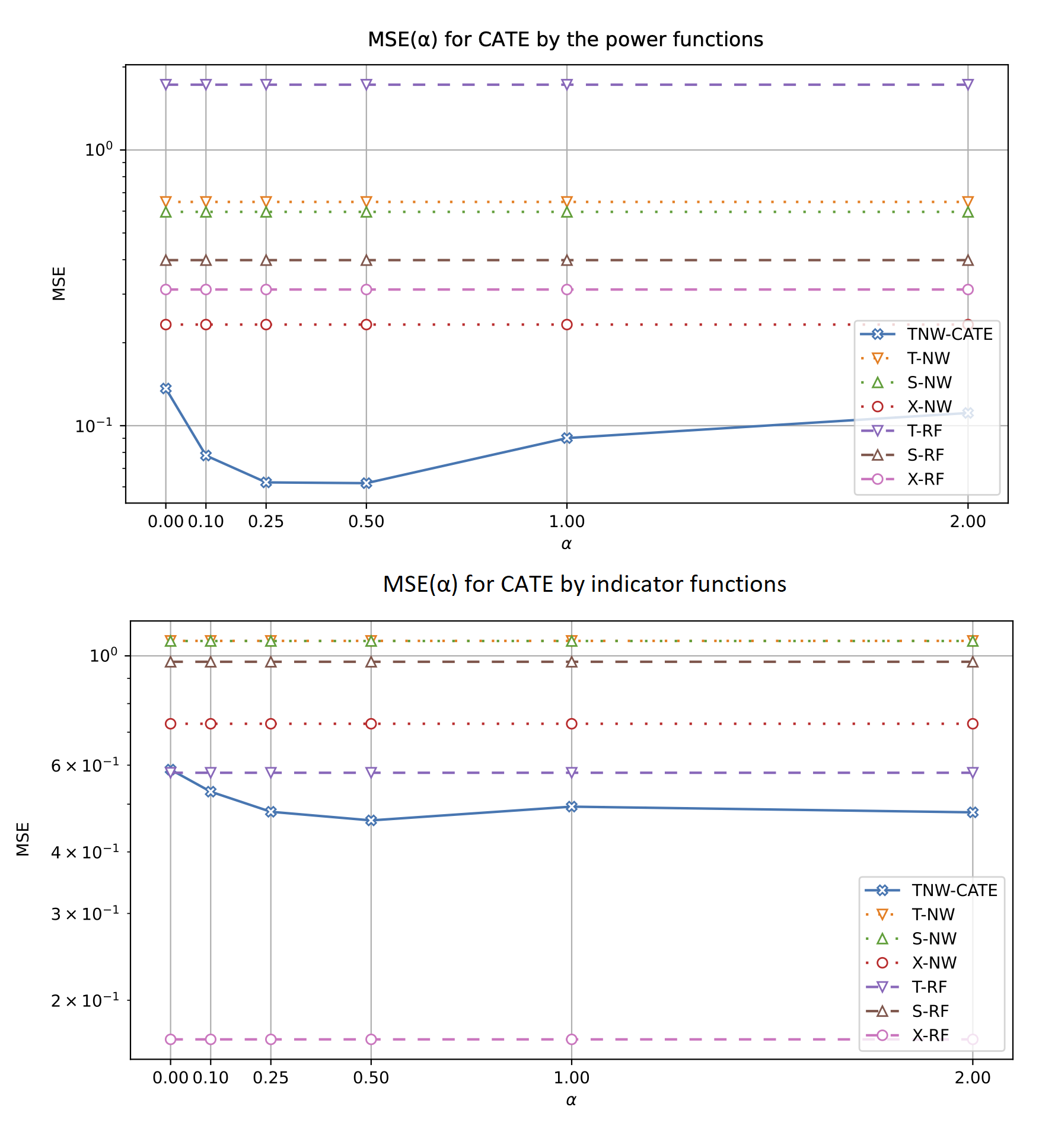}%
\caption{MSE of the CATE values as a function of the coefficient $\alpha$ when
power (the first plot) and indicator (the second plot) functions are used for
generating examples}%
\label{f:alpha_cate_pow_ind}%
\end{center}
\end{figure}

The next experiments allow us to investigate how the CATE estimators depend on
the value of hyperparameter $\alpha$ which controls the impact of the control
and treatment networks in the loss function (\ref{THTE_70}). The corresponding
numerical results are shown in Figs. \ref{f:alpha_cate_srl_log}%
-\ref{f:alpha_cate_pow_ind}. It should be noted that other models do not
depend on $\alpha$. One can see from Figs. \ref{f:alpha_cate_srl_log}%
-\ref{f:alpha_cate_pow_ind} that there is an optimal value of $\alpha$
minimizing MSE of TNW-CATE for every generating function. For example, the
optimal $\alpha$ for the spiral function is $0.1$. It can be seen from Fig.
\ref{f:alpha_cate_srl_log} that TNW-CATE can be inferior to other models when
$\alpha$ is not optimal. For example, the case $\alpha=0$ for the logarithmic
function leads to worse results of TNW-CATE in comparison with T-RF and S-RF.%

%

Numerical results are also presented in Table \ref{t:hte_1} where the MSE
values corresponding to different models by different generating functions are
given. The best results for every function are shown in bold. It can be seen
from Table \ref{t:hte_1} that TNW-CATE provides the best results for the
spiral, logarithmic and power functions. Moreover, the improvement is
sufficient. TNW-CATE is comparable with the X-NW and X-RF by the logarithmic
and power generating functions. At the same time, the proposed model is
inferior to X-RF for the indicator function.%

\begin{table}[tbp] \centering
\caption{The best MSE values of CATE for different models and by different generating functions}%
\begin{tabular}
[c]{ccccc}\hline
& \multicolumn{4}{c}{Functions}\\\hline
Model & Spiral & Logarithmic & Power & Indicator\\\hline
T-NW & $3.806$ & $0.377$ & $1.722$ & $0.650$\\\hline
S-NW & $3.629$ & $0.341$ & $1.719$ & $0.549$\\\hline
X-NW & $3.279$ & $0.542$ & $0.632$ & $0.353$\\\hline
T-RF & $2.278$ & $0.051$ & $2.743$ & $0.337$\\\hline
S-RF & $2.575$ & $0.060$ & $0.839$ & $0.434$\\\hline
X-RF & $1.385$ & $0.202$ & $0.805$ & $\mathbf{0.061}$\\\hline
TNW-CATE & $\mathbf{0.232}$ & $\mathbf{0.026}$ & $\mathbf{0.353}$ &
$0.257$\\\hline
\end{tabular}
\label{t:hte_1}%
\end{table}%

It is important to point out that application of many methods based on neural
networks, for example, DRNet \cite{Schwab-etal-2020}, DragonNet
\cite{Shi-Blei-Veitch-19}, FlexTENet \cite{Curth-Schaar-21} and VCNet
\cite{Nie-etal-21}, to comparing them with TNW-CATE have not been successful
because the aforementioned neural networks require the large amount of data
for training and the considered small datasets have led to the network
overfitting. Therefore, we considered models for comparison, which are based
on methods dealing with small data.

\section{Conclusion}

A new method (TNW-CATE) for solving the CATE problem has been studied. The
main idea behind TNW-CATE is to use the Nadaraya-Watson regression with
kernels which are implemented as neural networks of the considered specific
form and are trained on many randomly selected subsets of the treatment and
control data. With the proposed method, we aimed to avoid constructing the
regression function $g_{1}$ based only on the treatment group because it may
be small. Moreover, we aimed to avoid using standard kernels, for example,
Gaussian ones, in the Nadaraya-Watson regression. By training kernels on
controls (or controls and treatments), we aimed to transfer knowledge of the
feature vector structure in the control group to the treatment group.

In spite of the apparent complexity of the whole neural network for training,
TNW-CATE is actually simple because it can be realized as a single small
subnetwork implementing the kernel.

Numerical simulation experiments have illustrated the outperformance of
TNW-CATE for several datasets.

We used neural networks to learn kernels in the Nadaraya-Watson regression.
However, different models can be applied to the kernel implementation, for
example, the random forest \cite{Breiman-2001} or the gradient boosting
machine \cite{Friedman-2002}. The study of different models for estimating
CATE is a direction for further research. We also assumed the similarity
between structures of control and treatment data. This assumption can be
violated in some cases. Therefore, it is interesting to modify TNW-CATE to
take into account the violation. This is another direction for further
research. Another interesting direction for research is to incorporate robust
procedures and imprecise statistical models to deal with small datasets into
TNW-CATE. The incorporation of the models can provide estimates of CATE, which
are more robust than estimates obtained by means of TNW-CATE.

\bibliographystyle{unsrt}
\bibliography{Attention,Boosting,Deep_Forest,Explain,MYBIB,MYUSE,Survival_analysis,Transf_Learn,Treatment}

\end{document}